\documentclass{article}
\usepackage{arxiv}

\newif\ifieee
\ieeefalse

\usepackage{graphicx}
\usepackage{graphbox}
\usepackage{svg}
\usepackage{float}
\usepackage{subfig}
\usepackage{booktabs}
\usepackage{multirow}
\usepackage{multicol}
\usepackage[bottom]{footmisc}
\usepackage{siunitx}
\usepackage{tabulary}
\usepackage{amsmath}
\usepackage{mathtools}
\usepackage[export]{adjustbox}
\usepackage{todonotes}
\usepackage{soul}
\usepackage{xcolor}
\usepackage{cleveref}

\sethlcolor{green} 

\newcommand{\paperbibstyle}{IEEEtran}

\graphicspath{{figures/}}

\date{}

\newif\ifieee
\ieeefalse

\newif\ifanonymised
\anonymisedfalse

\begin{document}

\title{\LARGE \bf
Enabling Urgency-aware Robot Swarm Intralogistics \\ using Smart IoT Tags}



\ifieee

    \ifanonymised

        \author{Anonymous Authors}

    \else

        \author{
        Youssef Alboraei$^{1}$,
        Murray Groves$^{2}$,
        Shane Wen$^{2}$,
        Wenda Zhao$^{2}$,
        Senhui Qiu$^{2}$,
        Mohammud J. Bocus$^{2}$,\\
        Robert Piechocki$^{3}$,
        Sabine Hauert$^{1}$,
        and Kerstin Eder$^{2}$%
        \thanks{$^{1}$Y. Alboraei and S. Hauert are with the School of
        Engineering Mathematics and Technology and Bristol Robotics
        Laboratory, University of Bristol, UK
        (e-mail: youssef.alboraei@bristol.ac.uk;
        sabine.hauert@bristol.ac.uk).}%
        \thanks{$^{2}$M. Groves, S. Wen, W. Zhao, S. Qiu, M. J. Bocus,
        and K. Eder are with the School of Computer Science and
        Trustworthy Systems Laboratory, University of Bristol, UK
        (e-mail: murray.groves.2022@bristol.ac.uk;
        cc23766@alumni.bristol.ac.uk;
        ed20979@bristol.ac.uk;
        senhui.qiu@bristol.ac.uk;
        junaid.bocus@bristol.ac.uk;
        kerstin.eder@bristol.ac.uk).}%
        \thanks{$^{3}$R. Piechocki is with the School of Electrical,
        Electronic and Mechanical Engineering, University of Bristol, UK
        (e-mail: r.j.piechocki@bristol.ac.uk).}%
        }

    \fi

\else

\author{
    Youssef Alboraei \\
    School of Engineering Mathematics and Technology \\
    Bristol Robotics Laboratory \\
    University of Bristol, UK \\
    \texttt{youssef.alboraei@bristol.ac.uk}
    \And
    Murray Groves \\
    School of Computer Science \\
    Trustworthy Systems Laboratory \\
    University of Bristol, UK \\
    \texttt{murray.groves.2022@bristol.ac.uk}
    \And
    Shane Wen \\
    School of Computer Science \\
    Trustworthy Systems Laboratory \\
    University of Bristol, UK \\
    \texttt{cc23766@alumni.bristol.ac.uk}
    \AND
    Wenda Zhao \\
    School of Computer Science \\
    Trustworthy Systems Laboratory \\
    University of Bristol, UK \\
    \texttt{ed20979@bristol.ac.uk}
    \And
    Senhui Qiu \\
    School of Computer Science \\
    Trustworthy Systems Laboratory \\
    University of Bristol, UK \\
    \texttt{senhui.qiu@bristol.ac.uk}
    \And
    Mohammud J. Bocus \\
    School of Computer Science \\
    Trustworthy Systems Laboratory \\
    University of Bristol, UK \\
    \texttt{junaid.bocus@bristol.ac.uk}
    \AND
    Robert Piechocki \\
    School of Electrical, Electronic and Mechanical Engineering \\
    University of Bristol, UK \\
    \texttt{r.j.piechocki@bristol.ac.uk}
    \And
    Sabine Hauert \\
    School of Engineering Mathematics and Technology \\
    Bristol Robotics Laboratory \\
    University of Bristol, UK \\
    \texttt{sabine.hauert@bristol.ac.uk}
    \And
    Kerstin Eder \\
    School of Computer Science \\
    Trustworthy Systems Laboratory \\
    University of Bristol, UK \\
    \texttt{kerstin.eder@bristol.ac.uk}
}

\fi

\maketitle
\thispagestyle{empty}
\pagestyle{empty}

\providecommand{\hlrev}[1]{{\sethlcolor{cyan!45}\hl{#1}}}
\providecommand{\ci}[2]{\begin{tabular}{@{}c@{}}#1\\[-1.5pt]{\scriptsize #2}\end{tabular}}

\begin{abstract}
Warehouse items differ in how urgently they must be moved: perishable goods, pharmaceutical shipments, and just-in-time production materials must be delivered sooner than the rest of the stock. Decentralised robot swarms suit warehouses that cannot justify fixed automation infrastructure, but current swarm controllers treat all items alike or rely on an external scheduler to set priorities, so urgent items wait as long as ordinary ones. This paper presents a swarm logistics system in which each warehouse carrier holds an ultra-low-power Internet-of-Things (IoT) tag that broadcasts the urgency of its item over Bluetooth Low Energy (BLE). Robots read these broadcasts directly and weigh urgency against travel distance when choosing which carrier to serve, so prioritisation happens at the item level without central scheduling. The system is evaluated in simulation and validated on real robots and IoT-tagged carriers against a proximity-only baseline. In the physical trials, priority alignment (i.e. proportion of urgent items served first), improved from 0.41 to 0.64, 
with a non-significant trend toward lower 95th-percentile (P95) delivery latency and throughput within 1.2\% of the baseline. In simulation, the benefit grew with system size: across three larger configurations, P95 latency fell by 5.2\% to 11.8\% and priority alignment improved by 41.7\% to 51.6\%. Attaching urgency to the items themselves therefore allows a decentralised swarm to serve time-critical stock sooner while keeping the low infrastructure requirements that make swarm systems attractive for warehouse automation.
\end{abstract}

\section{INTRODUCTION}

Logistics and warehouse operations are facing increasing pressure to become more responsive, flexible, and cost-effective \cite{richeyResponsivenessViewLogistics2022}. While large-scale facilities can justify highly engineered automation stacks, dense sensing infrastructure, and tightly integrated warehouse management systems, many warehouses, distribution hubs, and back-room storage spaces remain either manual or only partially automated. In such settings, deploying fixed conveyors, centralised scheduling platforms, or dedicated localisation infrastructure may be economically unjustified, operationally disruptive, or insufficiently adaptable to changing layouts, inventories, and workloads. Infrastructure-light automation solutions that can be deployed rapidly, integrated with existing practices, and reconfigured with minimal setup effort are therefore needed \cite{dhaliwal2020rise}.

Swarm robotics is well suited to such settings. Swarm systems rely on large numbers of relatively simple robots that coordinate through local interactions rather than centralised control, providing robustness, scalability, and flexibility in uncertain or evolving environments~\cite{brambilla2013swarm,aguzzi2024engineering}. 
In logistics, these properties are attractive because they enable mobile and reconfigurable material-handling systems that can operate without extensive fixed infrastructure \cite{zhen2025optimizing} and can scale with operational demand \cite{hamannSwarmRoboticsFormal2018}. However, despite these advantages, effective coordination in logistics-oriented swarms remains challenging \cite{ikumapayi2024swarm}. Warehousing tasks are not merely spatial; they are also semantic and time-sensitive. Items may differ in urgency, handling requirements, or service priority, yet many decentralised swarm approaches treat tasks as largely homogeneous or assume that task priorities are assigned and managed externally.

This limitation motivates a shift from viewing warehouse items as passive objects to treating them as active information carriers. In this paper, we propose a decentralised logistics architecture in which item-specific semantics and urgency are embedded directly on carriers through smart tags. Rather than relying on a central server or fixed infrastructure to maintain global knowledge of task priorities, the proposed approach allows each tagged carrier to advertise its identity, urgency level, and status locally to nearby robots. As a result, robots can make context-aware, item-relevant decisions directly at the edge. The ultra-low-power IoT tag is central to this shift: unlike passive identifiers, it carries computation, state, and wireless communication on the item itself, which is what makes item-specific urgency available to robots without any supporting infrastructure, while keeping that information readable and adjustable by human operators.

To investigate this, we evaluate the proposed urgency-aware architecture in both simulation and physical experiments using Distributed Organisation and Transport System (DOTS) robots in a warehouse-inspired laboratory arena \cite{jones2022dots}. We further examine scalability in simulation by analysing how the decentralised coordination mechanism behaves as swarm size increases. Across these studies, we assess performance using logistics-relevant metrics including delivery latency, throughput, and priority responsiveness.

The main contributions of this paper are as follows:
\begin{itemize}
    \item We propose a decentralised swarm-logistics architecture in which smart IoT tags encode item urgency on carriers, together with a carrier-selection mechanism that combines this urgency with proximity in each robot's local decision, requiring no centralised scheduling.
    \item We evaluate the system in simulation and in physical experiments with real robots and IoT tags, showing that urgency-aware selection improves the prioritisation of urgent items relative to a proximity-only baseline while preserving throughput.
    \item We analyse scalability in simulation, showing that the benefits grow with swarm size while the throughput cost shrinks.
\end{itemize}

\begin{figure}
    \centering
    \includegraphics[width=0.9\linewidth]{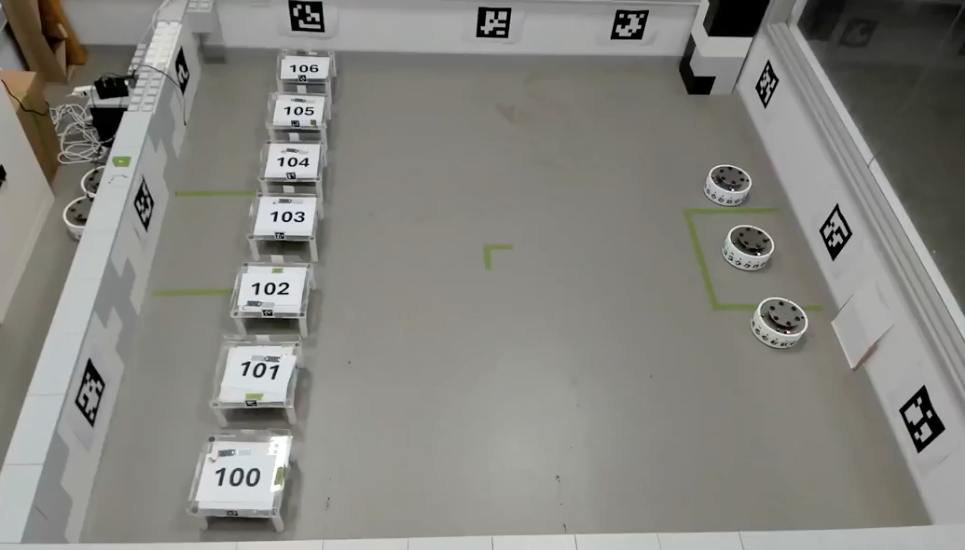}
    \caption{Swarm arena setup. 7 carriers arranged in pick-up zone (left), 3 DOTS robots start from drop-off zone (right). Drop-off zone identified by ArUco markers on right side.}
    \label{fig:arena}
\end{figure}
\section{Related Work}

Swarm robotics has attracted growing interest for warehouse automation because it enables robust, scalable, and flexible material handling through self-organisation and decentralised control \cite{brambilla2013swarm, sahinSwarmRoboticsSources2005}. Local interaction rather than centralised coordination lets swarms tolerate failures and adapt to dynamic environments with little fixed infrastructure.

A key challenge in such systems is distributed task allocation. Decentralised approaches allow robots to select tasks based on local sensing and limited communication, thereby preserving operation under partial failures or communication loss \cite{chenDecentralizedTaskPath2021, jones2020distributed}.
Market-based and auction-based methods have shown that competitive bidding can achieve efficient allocations while retaining decentralisation \cite{diasMarketBasedMultirobotCoordination2006, fazalTaskAllocationMultirobot2022}. More recently, methods that incorporate task importance or deadlines into the allocation objective have been shown to improve service for high-priority tasks \cite{notomista2019optimal, zhangDynamicPrioritizedTask2025}.

Low-power wireless communication is central to this capability: Bluetooth Low Energy (BLE) combines low energy consumption, moderate range, and reliable operation in congested 2.4~GHz industrial settings \cite{szyc2023bluetooth, OverviewEvaluationBluetooth}. Equipping items with communicating tags therefore enables robots to obtain task-relevant information locally, reducing reliance on centralised schedulers \cite{na2021bio, garnierBiologicalPrinciplesSwarm2007}.
Smart Internet-of-Things (IoT) tags extend this idea: unlike passive identifiers, they integrate sensing, computation, and wireless communication into physical assets, and can store and transmit information such as stock status, handling constraints, and deadlines \cite{khan2014design}. However, the use of smart tags as active coordinating elements in decentralised robotic logistics remains insufficiently explored.

Conventional warehouse automation often relies on centralised scheduling, which can provide efficient allocation but requires denser infrastructure and introduces single points of failure \cite{azadeh2019robotized}. In contrast, decentralised swarm approaches are inherently more robust and scalable, but they often neglect item-specific information, limiting responsiveness to time-sensitive items \cite{schranzSwarmRoboticBehaviors2020a}.
Existing decentralised methods typically represent urgency through heuristics such as shortest-distance-first or earliest-deadline-first selection \cite{ghassemiMultirobotTaskAllocation2022, choiConsensusBasedDecentralizedAuctions2009}, or through utility functions that balance travel cost and estimated task value \cite{kang2024optimization}. However, urgency is often assumed to be globally known, statically encoded, or centrally assigned, which restricts adaptation to changing item conditions.

Prior studies have shown that urgency-aware task allocation can improve service quality and reduce delays in multi-agent systems \cite{kalempaMultiRobotPreemptiveTask2021}. However, many of these studies remain simulation-based or assume continuous, high-bandwidth communication infrastructures \cite{kazim2025design}. Related work on event-driven and semantic communication in swarms further supports transmitting only the information that is relevant to collective behaviour \cite{anuraj2024dynamic, hickson2025beesics}.
Experimental evaluations that combine decentralised swarm coordination, ultra-low-power smart tags, and dynamically adjustable priorities in realistic logistics environments are still limited. 


\begin{figure}
    \centering
    \includegraphics[width=0.65\linewidth]{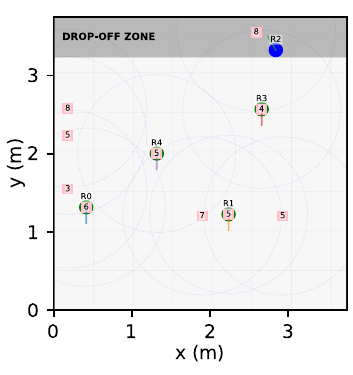}
    \caption{Snapshot from the 2D simulator, with axes in metres (the arena matches the 3.75$\times$3.75~m physical arena): robots (circles), carriers (squares) displaying their urgency value (0--9), and dotted circles marking each carrier's tag-detection range.
    }
    \label{fig:sim}
\end{figure}

\section{Methodology}

\subsection{Research Questions}
This study is guided by the following research questions:

\textbf{RQ1:} Does embedding item urgency directly into carriers improve the prioritisation of urgent items in decentralised swarm logistics, and reduce tail delivery latency (P95), while preserving overall throughput?

\textbf{RQ2:} How does the proposed urgency-aware coordination mechanism scale with increasing swarm size and workload relative to a proximity-based baseline?

\textbf{RQ3:} How does the urgency weight affect the trade-off
between prioritising urgent carriers and selecting carriers
closer to the robot in decentralised robot-carrier assignment?

\subsection{Experimental Scenario}
\begin{figure}[t]
    \centering
    \includegraphics[width=0.55\linewidth]{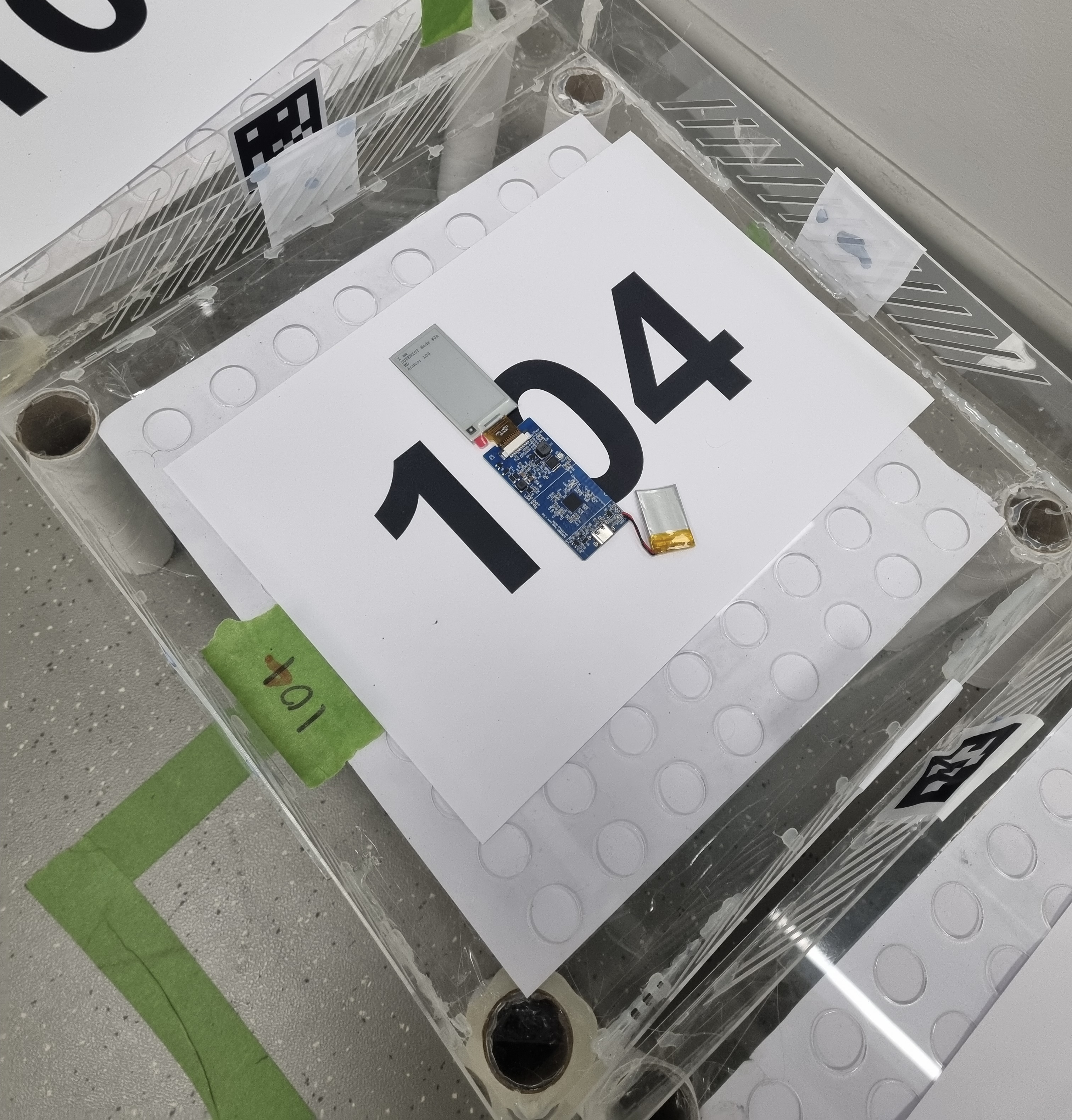}
    \caption{A carrier with its smart IoT tag. ArUco markers on the four sides (and underside) are used by robots for identification and docking. The tag broadcasts the carrier's ID and urgency level via BLE; the E-ink display shows the same information for human readability.}
    \label{fig:carrier}
\end{figure}

We consider an infrastructure-light warehouse scenario in which items are stored on four-legged carriers located in a pickup area (see left-hand side of Fig.~\ref{fig:arena}) and must be transported to a designated drop-off area (see right-hand side of Fig.~\ref{fig:arena}) by a decentralised robot swarm. Each carrier is equipped with a smart IoT tag that stores and broadcasts its identity, item-specific urgency, operational status and destination. The tag also shows the same information on an E-ink display, so that human operators can read and adjust item priorities at a glance, and each carrier additionally bears ArUco markers on its sides and underside that robots use to identify, approach and dock with it (Fig.~\ref{fig:carrier}).
Robots explore the environment independently and use locally available information to select, collect and deliver carriers without task assignments from a central scheduler. 

Two carrier-selection approaches are considered. Under the \textbf{proximity-only baseline}, robots select the nearest available carrier using its relative position, without considering the urgency information broadcast by its tag. Under the proposed \textbf{urgency-aware approach}, robots combine carrier proximity with the item-specific urgency advertised by the tag. The hardware, navigation behaviour, environment and remaining controller functions are common to both approaches, such that the principal difference is how carriers are selected.
A typical task cycle begins when a robot detects one or more available carriers during exploration. The robot obtains the carrier state and urgency from its BLE advertisement, selects a carrier according to the applicable policy, docks underneath it, activates its lifting mechanism and transports it to the drop-off area. The carrier status is updated throughout this process so that other robots can determine whether it remains available for collection.

\subsection{Controller Design}
\subsubsection{Robot Behaviour Controller}
Each robot operates autonomously using a modular behaviour-tree controller. The controller executes the following sequence:
\begin{itemize}
    \item \textbf{Random Walk:} move on a random heading until an obstacle or boundary is detected, then pick a new heading.
    \item \textbf{Select Carrier:} select an available carrier using the proximity-only or urgency-aware policy.
    \item \textbf{Predock:} move in front of the selected carrier and rotate to face it.
    \item \textbf{Dock and Centre:} drive underneath the carrier and align with its centre using the underside ArUco marker.
    \item \textbf{Transport and Release:} lift the carrier, transport it to the drop-off area, release it and resume exploration.
\end{itemize}
The tag state changes from \texttt{AWAITING PICKUP} to \texttt{PICKED UP} at collection and to \texttt{IDLE} on delivery, preventing other robots from selecting a carrier already being transported.

\setlength{\belowdisplayskip}{2pt} \setlength{\belowdisplayshortskip}{2pt}
\setlength{\abovedisplayskip}{0pt} \setlength{\abovedisplayshortskip}{0pt}
\subsubsection{Carrier-Selection Algorithm}
Each robot ranks candidate carriers using a score that combines the urgency broadcast by the tag with the carrier proximity. When urgency awareness is enabled, the score assigned by robot $r$ to carrier $c$ is defined as
\begin{equation}
s_{r,c}=u_c\left(\alpha+u_d(r,c)\right)
\label{eq:priority_score}
\end{equation}
where $u_c \in [0,9]$ denotes the carrier urgency level, $\alpha$ is a tunable urgency-weighting parameter, and $u_d(r,c)$ is a distance-dependent proximity term. The latter is given by
\begin{equation}
u_d(r,c)=9\left(1-\min\left(\frac{d_{r,c}}{3.6},\,1\right)\right)
\label{eq:distance_urgency}
\end{equation}
where $d_{r,c}=\left\|\mathbf{t}_{r,c}\right\|_2$ is the Euclidean distance to the carrier, with $\mathbf{t}_{r,c}$ the carrier position relative to robot $r$ in the robot-centred camera frame.

\begin{figure}[t]
    \centering    \includegraphics[width=0.6\linewidth]{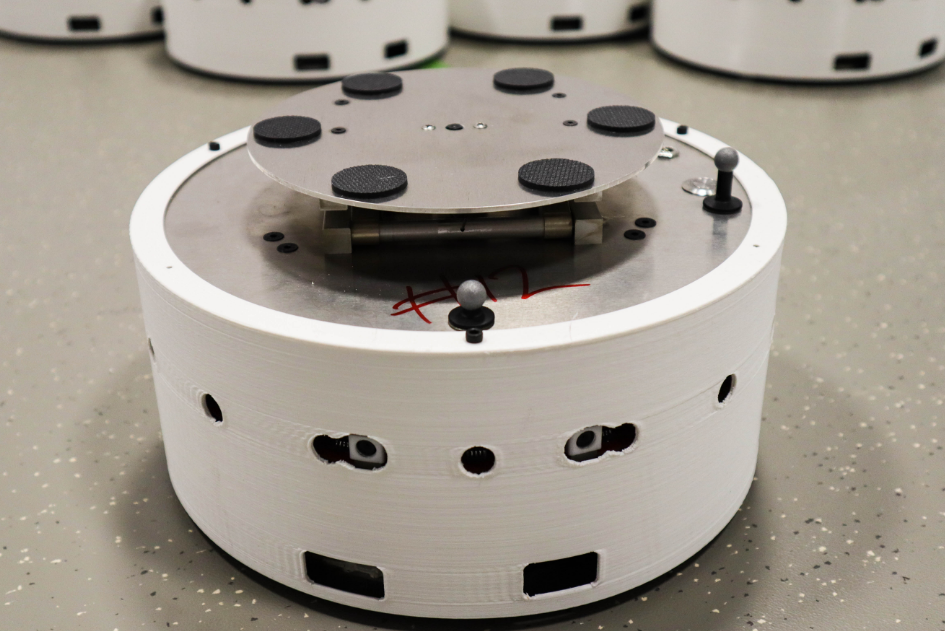}
    \caption{The DOTS robots use time-of-flight sensors for obstacle avoidance and cameras for ArUco markers processing to identify and localise the carriers and the drop-off zone.}
    \label{fig:dots}
\end{figure}
The proximity term in \eqref{eq:distance_urgency} is clipped and scaled to the range 0 to 9, corresponding to the urgency scale used by the smart tags. Nearby carriers therefore receive a larger proximity contribution, whereas carriers at distances greater than or equal to 3.6 m receive no additional proximity benefit. Representing urgency and proximity on the same numerical scale provides an interpretable basis for combining the two quantities.

The parameter, $\alpha$, controls the trade-off between urgency responsiveness and spatial efficiency. Increasing $\alpha$ gives greater influence to carrier urgency, whereas decreasing it gives greater influence to spatial proximity. Under urgency-aware operation, the robot selects the available carrier with the highest score in (1). Under the proximity-only baseline, urgency is ignored and the robot selects the carrier with the largest value of ($u_d(r,c)$), which is equivalent to selecting the nearest available carrier.

\subsection{Evaluation Metrics}
\label{sec:metrics}

System performance was evaluated across the following KPIs:

\begin{itemize}
    \item \textbf{P95 Latency:} The 95th percentile of per-carrier delivery latency, pooled across all deliveries of a condition. Each carrier's latency is measured relative to its target delivery time, which precedes the trial start by $(u_c/9)\cdot 240$~s, so that more urgent carriers accrue lateness sooner (Sec.~\ref{sec:results}).

    \item \textbf{Mean Latency:} The mean of the same per-carrier delivery latency across all deliveries of a condition.

    \item \textbf{Throughput:} The number of carriers delivered per 100 units of elapsed run time between a run's first and last delivery. Physical experiments use seconds, whereas simulations use simulator ticks. For fixed-duration scalability simulations, total deliveries provide an equivalent measure of throughput.
    
    \item \textbf{Priority Alignment (PA):} The extent to which high-urgency carriers are delivered before lower-urgency carriers. For an urgency threshold, $x$, this is calculated as the proportion of carriers within the top $x$\% by urgency that are also included within the first $x$\% of completed deliveries. Priority alignment is evaluated across thresholds from 5\% to 50\%.
\end{itemize}
The proximity-only (baseline) condition provides the reference for all comparisons. The same carrier and robot configuration is used for both approaches; however, the baseline controller ignores the urgency broadcast by the smart tags and selects carriers solely according to proximity.

\subsection{Software Environment}
\subsubsection{Robot Software}
The DOTS robots are controlled through Robot Operating System 2 (ROS~2) using behaviour trees; each robot executes an independent controller and selects carriers using locally available sensing and communication, with no global task list or centralised scheduler. A custom BLE library, running in its own thread, detects nearby tag advertisements and extracts carrier identity, urgency, status and destination, and camera-based ArUco processing provides carrier identification and relative localisation for docking \cite{jones2023frappe}.

\subsubsection{Simulation Environment}
A custom low-fidelity two-dimensional Python simulator was developed to represent the essential behaviour of the physical swarm-logistics system, as illustrated in Fig.~\ref{fig:sim}. The simulator models holonomic robot motion, robot-robot, robot-wall and robot-carrier collisions, random-walk exploration, range-limited tag detection and carrier pickup and delivery cycles.

The simulated robots execute the same proximity-only and urgency-aware carrier-selection rules as the physical robots. The simulator supports configurable numbers of robots and carriers, enabling experiments beyond the capacity of the physical platform.

\begin{figure}[t]
    \centering        \includegraphics[width=0.65\linewidth]{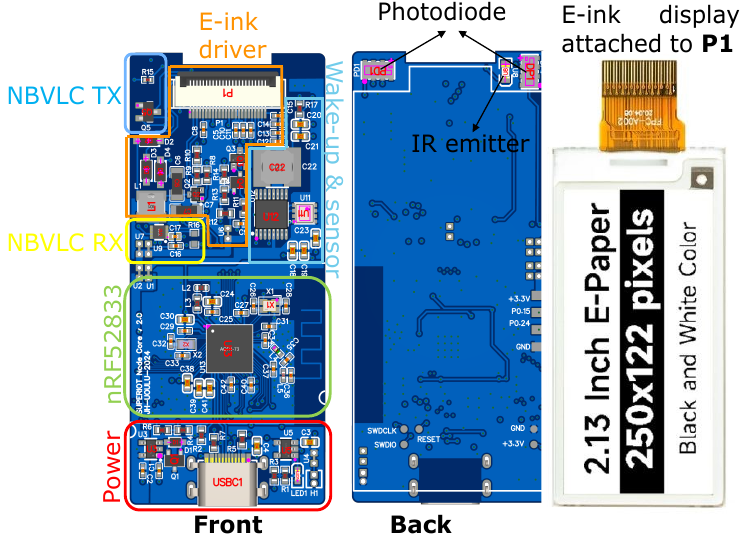}
    \caption{Custom-engineered reconfigurable IoT node.}
    \label{fig:node}
\end{figure}
\subsection{Hardware Platform}

\subsubsection{DOTS Robot Platform}
The Distributed Organisation and Transport System (DOTS) is an open-source industrial swarm-robotics platform designed for decentralised logistics research \cite{jones2022dots}. Each DOTS robot, shown in Fig.~\ref{fig:dots}, has a diameter of 250 mm and uses three omniwheels arranged at 120$^\circ$ intervals, providing holonomic motion and rapid omnidirectional movement at speeds of up to 2 m/s. Its lifting mechanism supports payloads of up to 2 kg.

Onboard computation (a six-core ARM single-board computer with GPU) supports local perception, decision-making and control. Each robot carries four perimeter cameras, an upward-facing camera, 16 infrared time-of-flight sensors, Wi-Fi, two BLE~5 radios and an ultra-wideband radio. In the present study, the time-of-flight sensors support obstacle avoidance, and the cameras process ArUco markers to identify and localise carriers and the drop-off area (Fig.~\ref{fig:arena}).

\subsubsection{Custom Smart IoT Tags}
The reconfigurable IoT node shown in Fig.~\ref{fig:node} was custom designed and implemented within the SUPERIOT project as a prototype of its sustainable and reconfigurable IoT-node concept \cite{superiot}. SUPERIOT investigates IoT systems combining radio and optical connectivity, low-power operation, sensing and adaptable node functionality.

The node is centred on an nRF52833 BLE system-on-chip and additionally carries a visible-light-communication transceiver, an environmental sensor, a light-based wake-up circuit and a 2.13-inch E-ink display; only the BLE and E-ink functions are used in this study.

The node broadcasts its state through BLE advertisements containing manufacturer-specific data. Each advertisement includes the following six-byte payload: \texttt{[0xFF, 0xFF, urgency, status, destination, ArUco ID]}. 
Urgency is encoded on a scale from 0 to 9. The status field indicates whether a carrier is \texttt{IDLE (0x00)}, \texttt{AWAITING PICKUP (0x01)} or \texttt{PICKED UP (0x02)}. The remaining fields encode destination and visual-identifier information. This compact representation allows nearby robots to obtain item-specific priority and state information directly from periodic advertisements. The node state can also be updated remotely through a BLE Generic Attribute Profile (GATT) characteristic, after which the advertising payload is refreshed. This write path is what makes item priorities dynamically adjustable: an operator or warehouse management system can re-prioritise an item mid-operation (e.g., when an order becomes urgent), and the experiment infrastructure uses the same mechanism to record status transitions at pickup and delivery.

\subsubsection{Physical Environment}

Physical experiments were conducted in a 3.75~m~$\times$~3.75~m arena representing a simplified warehouse environment with designated pickup and drop-off areas, as shown in Fig.~\ref{fig:arena}. 
Three DOTS robots operate as mobile transport agents, while
seven carriers, each measuring
355~mm~$\times$~355~mm,
serve as proxies for warehouse items.

Each carrier is fitted with a custom smart IoT tag and ArUco markers on its sides and underside (Fig.~\ref{fig:carrier}); ArUco markers also identify the drop-off area. The same environment and robot-control functions are used for both conditions.

\begin{figure}[t]
\centering
\includegraphics[width=0.95\linewidth]{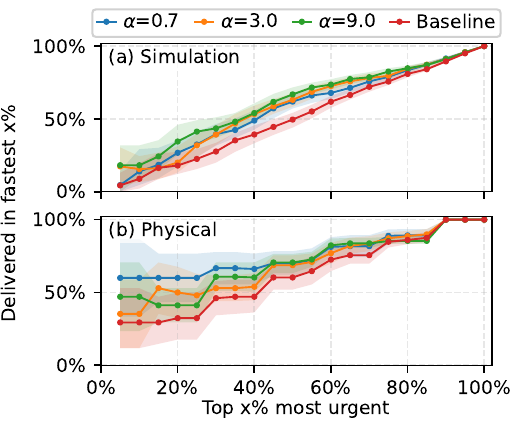}
\caption{Priority alignment as a function of the most urgent carrier subset for (a) simulation-based controller selection using three robots and seven carriers (100--101 runs per condition) and (b) physical validation using three robots and seven carriers (15--17 valid runs per condition). Curves show the mean across runs, and shaded bands show 95\% bootstrap confidence intervals (trial resampling). Higher values indicate stronger alignment between robot assignments and carrier urgency. Urgency-aware assignment improves priority alignment in both environments, with $\alpha=0.7$ providing the strongest improvement on the physical platform.}
\label{fig:PA}
\end{figure}

\begin{figure}[t]
\centering
\includegraphics[width=0.95\linewidth]{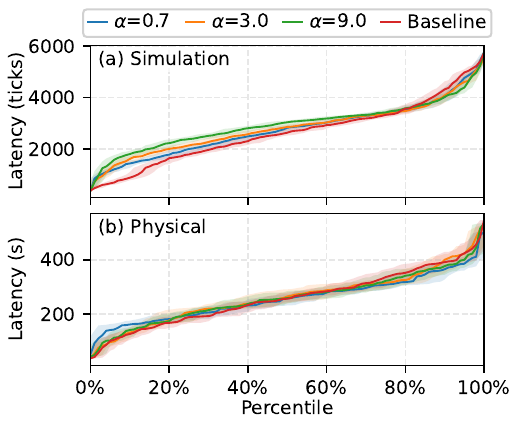}
\caption{Delivery-latency distributions for (a) simulation-based controller selection (100--101 runs per condition) and (b) physical validation (15--17 valid runs per condition). Curves show the mean percentile latency, and shaded bands show 95\% bootstrap confidence intervals (trial resampling). Lower latency is preferable. Urgency-aware assignment shows a consistent trend toward lower upper-tail latency, although the most favourable urgency weight differs between the simulated and physical environments.}
\label{fig:latency}
\end{figure}

\section{Results }\label{sec:results}

\subsection{Experimental Configurations and Analysis}
The experiments were conducted in three stages: simulation-based controller selection, physical validation, and simulation-based scalability analysis. Results are presented with the two simulation studies first, followed by the physical validation.

First, the influence of the urgency-weighting parameter ($\alpha$) in the controller score defined in Eq~(\ref{eq:priority_score}) was investigated in simulation. Intuitively, increasing $\alpha$ gives greater importance to carrier urgency relative to robot–carrier proximity. Candidate values were evaluated using a configuration of three robots and seven carriers. The urgency--distance balance was explored in two stages. A preliminary sweep varied the relative weight of urgency against distance from 0 to 1 in increments of 0.1, identifying mild urgency weighting as the most promising regime. Candidate weightings from this region (0.3, 0.5, 0.7 and 0.9) were then evaluated as settings of $\alpha$ under the scoring rule of \eqref{eq:priority_score}, with $\alpha$=0.7 outperforming the neighbouring setting $\alpha$=0.9 in both tail latency and completed deliveries. It was therefore carried forward, together with $\alpha$=3.0 and $\alpha$=9.0 as representative moderate and strong weightings on the 0--9 urgency scale.
The simulated robots moved at a maximum speed of 0.5 m/s within a 3.75 m $\times$ 3.75 m arena.
Each simulated trial ran for a fixed duration of 3{,}000 simulator ticks. Simulated carriers were initialised with waiting times sampled uniformly over this window, and carrier urgency grew linearly with waiting time up to the maximum value of 9. In the physical experiments, required delivery times were instead sampled uniformly between 0 and 240~s and mapped to static integer urgency values from 0 to 9, with shorter required delivery times corresponding to higher urgency. These three conditions were compared with the proximity-based baseline controller, which did not use carrier urgency. Each of the four retained conditions was evaluated over 100--101 simulation runs with controlled random seeds.

The selected controller settings were subsequently evaluated on the physical platform using three robots and seven carriers. Each of the four controller conditions (baseline, $\alpha$=0.7, $\alpha$=3.0, and $\alpha$=9.0) was evaluated in 15 to 17 valid runs (66 in total); a small number of additional runs were discarded due to hardware faults such as docking accidents.
The number of runs per condition was set by the testbed time available for the study.
At the beginning of each run, the carriers were positioned in the pickup area and the robots in the drop-off area. Carrier positions and urgency values were rearranged between runs using controlled randomisation. The robot speed was limited to 0.5 m/s. The system was fully reset between runs, including tag reinitialization and the relaunching of the robot controllers and BLE communication. Performance was recorded using both visual observations and ROS~2 logs. Tags remained present in the baseline condition, but their urgency information was ignored by the controller. All other experimental parameters were held constant.

Finally, scalability was evaluated in simulation using configurations of 5 robots and 10 carriers, 10 robots and 20 carriers, and 20 robots and 40 carriers. For each configuration, the baseline controller was compared with the mild urgency-aware controller, $\alpha$=0.7, over 21 runs per condition using controlled random seeds, with fixed trial durations of 3{,}000, 6{,}000 and 12{,}000 simulator ticks for the three scales, respectively.

\begin{figure}[t]
    \centering
    \includegraphics[width=0.8\linewidth]{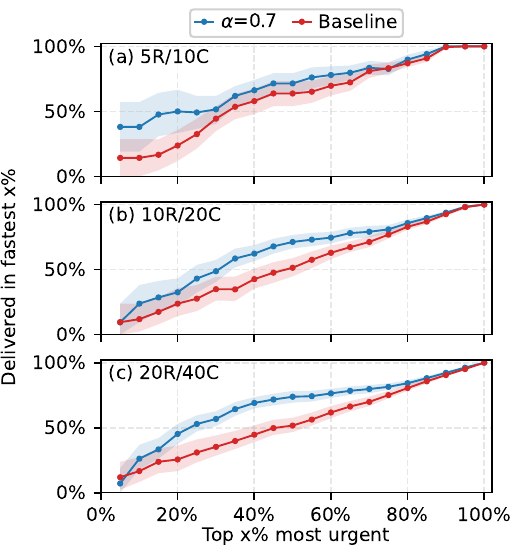}
    \caption{Priority alignment for the baseline and urgency-aware ($\alpha$=0.7) controllers at three simulated scales: (a) 5R/10C, (b) 10R/20C, and (c) 20R/40C, where $R$ and $C$ denote robots and carriers, respectively. Curves show the mean across 21 runs per condition, and shaded bands show 95\% bootstrap confidence intervals (trial resampling); panels are labelled by scale only and axes are reported as in Fig.~\ref{fig:PA}. The improvement produced by urgency-aware assignment becomes more pronounced as system size increases.}
    \label{fig:priority-stacked}
\end{figure}

Tables~\labelcref{tab:sim-data,tab:exp-data,tab:scaled-envs} retain the principal numerical results, while Figs.~\labelcref{fig:PA,fig:latency,fig:priority-stacked} illustrate the corresponding distributions and trends. The figure curves represent means across repeated runs, with shaded bands showing 95\% bootstrap confidence intervals (CIs).
The following analysis focuses on the direction, magnitude, and consistency of the observed effects rather than individual experimental observations.

\begin{table}[t]
    \centering
\caption{Simulation results for controller selection using three robots and seven carriers. Values summarize 100--101 runs per controller condition; brackets give 95\% bootstrap confidence intervals (trial resampling).}
\label{tab:sim-data}
\resizebox{\columnwidth}{!}{%
    \begin{tabular}{|c|c|c|c|c|}
        \hline
        {\textbf{Metric}} & {\textbf{Baseline}} & {\textbf{$\alpha$=0.7}} & {\textbf{$\alpha$=3.0}} & {\textbf{$\alpha$=9.0}} \\
        \hline
        P95 Latency (ticks)  & \ci{4979}{[4640, 5086]} & \ci{4765}{[4390, 4913]} & \ci{4608}{[4385, 4912]} & \ci{4390}{[4164, 4733]} \\
        \hline
        Mean Latency (ticks) & \ci{2638}{[2531, 2738]} & \ci{2767}{[2663, 2869]} & \ci{2820}{[2721, 2918]} & \ci{2947}{[2853, 3037]} \\
        \hline
        Throughput (/100 ticks)   & \ci{0.266}{[0.250, 0.285]} & \ci{0.251}{[0.235, 0.268]} & \ci{0.276}{[0.258, 0.296]} & \ci{0.258}{[0.243, 0.275]} \\
        \hline
        PA (5--50\%)   & \ci{0.27}{[0.22, 0.31]} & \ci{0.35}{[0.30, 0.40]} & \ci{0.36}{[0.31, 0.42]} & \ci{0.41}{[0.36, 0.47]} \\
        \hline
    \end{tabular}
    }
\end{table}


\subsection{Simulation-Based Controller Selection: The Urgency--Distance Trade-off (RQ3)}
The simulation experiments show that incorporating urgency into the robot–carrier assignment policy improves prioritization of urgent carriers and reduces tail delivery latency. However, increasing the urgency weight also introduces a trade-off between prioritization and average system performance.

As shown in Table~\ref{tab:sim-data} and Fig.~\ref{fig:PA}(a), each urgency-aware condition produced greater priority alignment than the baseline. The improvement was most pronounced for $\alpha$=9.0, indicating that stronger urgency weighting increased the likelihood that robots selected carriers from the most urgent subset.

A similar trend is visible in the upper tail of the latency distributions in Fig.~\ref{fig:latency}(a). P95 delivery latency decreased as $\alpha$ increased, showing that urgency-aware assignment particularly benefited carriers that would otherwise experience long delivery delays. However, mean latency increased with stronger urgency weighting. This indicates that prioritizing urgent carriers can delay less urgent carriers, even while improving the upper tail of the latency distribution.

Throughput did not change monotonically with $\alpha$. The moderate urgency setting, $\alpha$=3.0, produced the highest simulated throughput, whereas the mild and strong settings produced slightly lower throughput than the baseline. The three selected values were retained for physical evaluation because they represent distinct levels of urgency sensitivity and expose the trade-off between priority alignment, tail latency, and overall efficiency.

In terms of statistical reliability, the priority-alignment gains in simulation are significant for all three weightings (bootstrap $p=0.03$, $p<0.01$, and $p<0.001$ for $\alpha=0.7$, $3.0$, $9.0$, respectively), whereas the P95 latency reduction reaches significance only at the strongest weighting ($\alpha=9.0$: $-514$ ticks, 95\% CI $[-835, -80]$, $p=0.02$).

\textbf{Answering RQ3:} the exploration of $\alpha$ shows a trade-off between urgency responsiveness and spatial efficiency: moderate weighting reduces tail latency without degrading mean performance, and the non-monotonic relationship means urgency must be balanced against proximity.

\begin{table}[t]
\caption{Scalability results comparing the baseline and urgency-aware ($\alpha$=0.7) controllers. Each configuration was evaluated over 21 simulation runs per condition. $R$ and $C$ denote robots and carriers, respectively. Positive relative changes indicate improvement in the desired direction, including reductions in latency. Brackets give 95\% bootstrap confidence intervals (trial resampling); $^{*}$ marks changes significant at $p<0.05$ (two-sided trial-level bootstrap).}
\centering
\resizebox{\columnwidth}{!}{%
\begin{tabular}{|c|c|c|c|c|}
\hline
\textbf{Scale} & \textbf{Metric} & \textbf{Baseline} & \textbf{$\alpha$=0.7} & \textbf{$\Delta$} \\
\hline
\multirow{3}{*}{5R/10C} &
Total Deliveries & \ci{160}{[145, 174]} & \ci{145}{[129, 160]} & $-9.4\%$ \\
\cline{2-5}
& P95 Lat. (ticks) & \ci{4{,}441}{[3{,}981, 4{,}928]} & \ci{4{,}208}{[3{,}633, 4{,}454]} & $+5.2\%$ \\
\cline{2-5}
& PA (5--50\%) & \ci{0.39}{[0.30, 0.48]} & \ci{0.55}{[0.46, 0.63]} & $+41.7\%^{*}$ \\
\hline
\multirow{3}{*}{10R/20C} &
Total Deliveries & \ci{394}{[378, 407]} & \ci{394}{[381, 405]} & $+0.0\%$ \\
\cline{2-5}
& P95 Lat. (ticks) & \ci{8{,}790}{[8{,}301, 9{,}279]} & \ci{7{,}893}{[7{,}492, 8{,}535]} & $+10.2\%^{*}$ \\
\cline{2-5}
& PA (5--50\%) & \ci{0.30}{[0.23, 0.39]} & \ci{0.45}{[0.38, 0.52]} & $+47.9\%^{*}$ \\
\hline
\multirow{3}{*}{20R/40C} &
Total Deliveries & \ci{840}{[840, 840]} & \ci{834}{[830, 838]} & $-0.7\%^{*}$ \\
\cline{2-5}
& P95 Lat. (ticks) & \ci{14{,}097}{[13{,}705, 14{,}889]} & \ci{12{,}428}{[12{,}180, 12{,}771]} & $+11.8\%^{*}$ \\
\cline{2-5}
& PA (5--50\%) & \ci{0.33}{[0.27, 0.41]} & \ci{0.50}{[0.45, 0.56]} & $+51.6\%^{*}$ \\
\hline
\end{tabular}
}
\label{tab:scaled-envs}
\end{table}

\subsection{Simulation-Based Scalability Analysis (RQ2)}
The principal result of the scalability analysis is that the benefit of urgency-aware assignment becomes more evident as the numbers of robots and carriers increase. In particular, improvements in priority alignment and P95 latency grew with system size, while the associated throughput penalty decreased.

Before evaluating scalability, the simulation baseline was compared qualitatively with the physical baseline reported in Sec.~\ref{subsec:physical}: their absolute values lie on different time scales, but both produced similar latency-distribution and priority-alignment behaviour, supporting the use of the simulator for trend analysis rather than exact prediction.

As summarised in Table~\ref{tab:scaled-envs}, the reduction in P95 latency increased from 5.2\% in the smallest configuration to 11.8\% in the largest configuration. Priority alignment improved by between 41.7\% and 51.6\%, with the greatest relative improvement occurring in the largest system. The priority-alignment improvement is statistically significant at all three scales (trial-level bootstrap $p=0.013$, $p=0.012$ and $p=0.001$), and the P95 reduction is significant at the two larger scales ($p=0.023$ and $p<0.001$) but not at 5R/10C ($p=0.45$).

The effect on throughput (total deliveries) also became less pronounced with scale. The smallest configuration exhibited a 9.4\% reduction in completed deliveries (not statistically significant, $p=0.19$), whereas the difference was negligible for the two larger configurations. Mean and median latency likewise remained close to the baseline at the larger scales. These results indicate that larger systems provide more opportunities for the controller to assign nearby robots to urgent carriers without substantially reducing the service available to other carriers.

\textbf{Answering RQ2:} the benefit increases as the system grows. Priority-alignment gains exceed 40\% and are significant at every scale, and P95 latency improvements progress from 5.2\% to 11.8\%, reaching significance at the two larger scales. The throughput penalty observed at the smallest scale ($-9.4$\% at 5R/10C, not significant) vanishes at the larger scales, indicating that the cost of prioritising urgent carriers is amortised as the swarm grows.

\subsection{Physical Validation: Significant Priority-Alignment Gains Without Throughput Cost (RQ1)}
\label{subsec:physical}
The physical experiments reproduced the principal trend observed in simulation: incorporating urgency improved priority alignment and could reduce tail delivery latency without materially changing mean delivery latency. However, the mild urgency setting, rather than the strongest setting, produced the most favourable overall balance on the physical platform.

\begin{table}[t]
    \centering
\caption{Physical experimental results using three robots and seven carriers. Values summarize 15--17 valid runs per controller condition; brackets give 95\% bootstrap confidence intervals (trial resampling).}
\label{tab:exp-data}
\resizebox{\columnwidth}{!}{%
    \begin{tabular}{|c|c|c|c|c|}
        \hline
        \textbf{Metric}       & \textbf{Baseline} & \textbf{$\alpha$=0.7} & \textbf{$\alpha$=3.0} & \textbf{$\alpha$=9.0} \\
        \hline
        P95 Latency (s)  & \ci{424}{[387, 462]} & \ci{386}{[361, 411]} & \ci{422}{[401, 453]} & \ci{400}{[372, 434]} \\
        \hline
        Mean Latency (s) & \ci{260}{[244, 278]} & \ci{258}{[243, 273]} & \ci{263}{[246, 282]} & \ci{261}{[243, 279]} \\
        \hline
        Throughput (/100 s)   & \ci{3.84}{[3.30, 4.49]} & \ci{3.79}{[3.38, 4.27]} & \ci{3.48}{[2.99, 4.05]} & \ci{3.60}{[3.15, 4.14]} \\
        \hline
        PA (5--50\%)   & \ci{0.41}{[0.30, 0.54]} & \ci{0.64}{[0.54, 0.74]} & \ci{0.52}{[0.41, 0.63]} & \ci{0.54}{[0.46, 0.62]} \\
        \hline
    \end{tabular}
    }
\end{table}


As shown in Table~\ref{tab:exp-data} and Fig.~\ref{fig:PA}(b), all three urgency-aware settings increased priority alignment relative to the baseline. The largest improvement was obtained with $\alpha$=0.7, for which priority alignment increased from 0.41 to 0.64; this improvement is statistically significant ($+0.22$, 95\% CI $[0.06, 0.38]$, bootstrap $p<0.01$; Mann--Whitney on per-trial values $p=0.015$). This result demonstrates that even mild urgency weighting can substantially improve the controller's ability to prioritize urgent carriers.

The mild urgency setting also showed a trend toward lower P95 delivery latency (a 9.1\% reduction, from 424~s to 386~s), although this difference is not statistically significant at the achieved sample size (bootstrap difference $-31$~s, 95\% CI $[-81, +16]$, $p=0.24$); mean latency and throughput remained close to their baseline values (throughput $-1.2$\%, n.s.). The higher urgency weights produced smaller, non-significant improvements in P95 latency and larger throughput penalties. These results indicate that excessively strong urgency weighting can impair overall coordination by directing several robots towards highly urgent carriers while giving insufficient consideration to spatial efficiency.

The physical and simulation experiments therefore agree on the main effect: urgency-aware assignment improves priority alignment and can reduce tail latency, but the physical platform favoured a lower $\alpha$, likely because of effects simplified or absent in the simulator (robot-robot interactions, localisation and communication uncertainty, and congestion around urgent carriers). The results support $\alpha$=0.7 as the most appropriate setting for the tested physical configuration.

\textbf{Answering RQ1:} physical experiments show that urgency-aware coordination significantly improves the order in which urgent items are served (priority alignment $+0.22$, $p<0.01$ at $\alpha=0.7$) while maintaining throughput within 1.2\% of baseline. Tail latency improved by 9.1\% in aggregate, a trend consistent with the simulation results, though not individually significant at 15--17 trials per condition; confirming the size of this effect will require larger-scale trials.

\section{Discussion}
This work demonstrates that embedding urgency information within carriers through smart IoT tags enables a robot swarm to respond more effectively to time-sensitive intralogistics tasks while retaining decentralised coordination.

\subsection{Implications and Future Opportunities}
The proposed approach extends decentralised proximity-based coordination by incorporating item-specific urgency into each robot's local carrier-selection decision, requiring only a modification to the selection score while the broader coordination architecture remains unchanged.

Future work will evaluate the system in larger warehouse environments with increased numbers of physical robots and carriers. It will also investigate live updates to the information stored on the tags and adaptive tuning of the urgency weight according to system and environmental conditions.

\section{Conclusion}
This paper presents the design, implementation, and experimental validation of an urgency-aware swarm logistics system integrating smart IoT tags with decentralised DOTS robots. Carriers are fitted with ultra-low-power IoT nodes that broadcast item urgency via BLE, so that priority information travels with the items themselves rather than through a central scheduler.

Physical trials showed a significant improvement in the prioritisation of urgent items, together with a trend toward lower tail latency, while maintaining throughput close to the baseline.
Simulation-based scalability assessments revealed consistent reductions in tail latency and improvements in priority alignment across the evaluated scales. The throughput trade-off diminished as the system grew, demonstrating that urgency-driven coordination can enhance service quality with limited impact on overall output.
The approach requires no fixed infrastructure, which makes it suitable for warehouses where conventional automation is not justified.

\ifanonymised
\else
\section*{Acknowledgements}
This work was supported by the Engineering and Physical Sciences Research Council (EPSRC) through the Co-AIMS project [UKRI1890].
This work was also conducted as part of the SUPERIOT project.

The SUPERIOT project has received funding from the Smart Networks and Services Joint Undertaking (SNS JU) under the European Union’s Horizon Europe research and innovation programme under Grant Agreement No 101096021, including funding under the UK government’s Horizon Europe funding guarantee, UKRI Grant Reference Number 10053751. 

Views and opinions expressed are however those of the authors only and do not necessarily reflect those of the European Union, SNS JU or UKRI. The European Union, SNS JU or UKRI cannot be held responsible for them.
\fi

\bibliographystyle{\paperbibstyle}
{\footnotesize
\bibliography{ref.bib}}

@article{schranzSwarmRoboticBehaviors2020a,
  title = {Swarm {{Robotic Behaviors}} and {{Current Applications}}},
  author = {Schranz, Melanie and Umlauft, Martina and Sende, Micha and Elmenreich, Wilfried},
  journal = {Frontiers in Robotics and AI},
  volume = {7},
  publisher = {Frontiers},
  issn = {2296-9144},
  doi = {10.3389/frobt.2020.00036},
  url = {https://www.frontiersin.org/journals/robotics-and-ai/articles/10.3389/frobt.2020.00036/full},
  year = {2020},
  month = apr
}

@article{choiConsensusBasedDecentralizedAuctions2009,
  title = {Consensus-{{Based Decentralized Auctions}} for {{Robust Task Allocation}}},
  author = {Choi, Han-Lim and Brunet, Luc and How, Jonathan P.},
  journal = {IEEE Transactions on Robotics},
  volume = {25},
  number = {4},
  pages = {912--926},
  issn = {1941-0468},
  doi = {10.1109/TRO.2009.2022423},
  url = {https://ieeexplore.ieee.org/document/5072249},
  year = {2009},
  month = aug
}

@article{ghassemiMultirobotTaskAllocation2022,
  title = {Multi-Robot Task Allocation in Disaster Response: {{Addressing}} Dynamic Tasks with Deadlines and Robots with Range and Payload Constraints},
  author = {Ghassemi, Payam and Chowdhury, Souma},
  journal = {Robotics and Autonomous Systems},
  volume = {147},
  pages = {103905},
  issn = {0921-8890},
  doi = {10.1016/j.robot.2021.103905},
  url = {https://www.sciencedirect.com/science/article/pii/S0921889021001901},
  year = {2022},
  month = jan
}

@article{kalempaMultiRobotPreemptiveTask2021,
  title = {Multi-{{Robot Preemptive Task Scheduling}} with {{Fault Recovery}}: {{A Novel Approach}} to {{Automatic Logistics}} of {{Smart Factories}}},
  author = {Kalempa, Vivian Cremer and Piardi, Luis and Limeira, Marcelo and de Oliveira, Andr{\'e} Schneider},
  journal = {Sensors},
  volume = {21},
  number = {19},
  pages = {6536},
  publisher = {Multidisciplinary Digital Publishing Institute},
  issn = {1424-8220},
  doi = {10.3390/s21196536},
  url = {https://www.mdpi.com/1424-8220/21/19/6536},
  year = {2021},
  month = jan
}

@article{richeyResponsivenessViewLogistics2022,
  title = {A {{Responsiveness View}} of Logistics and Supply Chain Management},
  author = {Richey, Robert Glenn and Roath, Anthony S. and Adams, Frank G. and Wieland, Andreas},
  journal = {Journal of Business Logistics},
  volume = {43},
  number = {1},
  pages = {62--91},
  issn = {2158-1592},
  doi = {10.1111/jbl.12290},
  url = {https://onlinelibrary.wiley.com/doi/abs/10.1111/jbl.12290},
  year = {2022}
}

@article{chenDecentralizedTaskPath2021,
  title = {Decentralized {{Task}} and {{Path Planning}} for {{Multi-Robot Systems}}},
  author = {Chen, Yuxiao and Rosolia, Ugo and Ames, Aaron D.},
  year = 2021,
  month = jul,
  journal = {IEEE Robotics and Automation Letters},
  volume = {6},
  number = {3},
  pages = {4337--4344},
  issn = {2377-3766, 2377-3774},
  doi = {10.1109/LRA.2021.3068103}
}

@article{diasMarketBasedMultirobotCoordination2006,
  title = {Market-{{Based Multirobot Coordination}}: {{A Survey}} and {{Analysis}}},
  author = {Dias, M.B. and Zlot, R. and Kalra, N. and Stentz, A.},
  journal = {Proceedings of the IEEE},
  volume = {94},
  number = {7},
  pages = {1257--1270},
  issn = {1558-2256},
  doi = {10.1109/JPROC.2006.876939},
  url = {https://ieeexplore.ieee.org/document/1677943},
  year = {2006},
  month = jul
}

@article{fazalTaskAllocationMultirobot2022,
  title = {Task Allocation in Multi-Robot System Using Resource Sharing with Dynamic Threshold Approach},
  author = {Fazal, Nayyer and Khan, Muhammad Tahir and Anwar, Shahzad and Iqbal, Javaid and Khan, Shahbaz},
  year = 2022,
  month = may,
  journal = {PLoS ONE},
  volume = {17},
  number = {5},
  pages = {e0267982},
  issn = {1932-6203},
  doi = {10.1371/journal.pone.0267982},
  pmcid = {PMC9067701},
  pmid = {35507628}
}

@article{garnierBiologicalPrinciplesSwarm2007,
  title = {The Biological Principles of Swarm Intelligence},
  author = {Garnier, Simon and Gautrais, Jacques and Theraulaz, Guy},
  year = 2007,
  month = jun,
  journal = {Swarm Intelligence},
  volume = {1},
  number = {1},
  pages = {3--31},
  issn = {1935-3820},
  doi = {10.1007/s11721-007-0004-y},
  langid = {english},
}

@book{hamannSwarmRoboticsFormal2018,
  title = {Swarm {{Robotics}}: {{A Formal Approach}}},
  shorttitle = {Swarm {{Robotics}}},
  author = {Hamann, Heiko},
  year = 2018,
  publisher = {Springer International Publishing},
  address = {Cham},
  doi = {10.1007/978-3-319-74528-2},
  copyright = {http://www.springer.com/tdm},
  isbn = {978-3-319-74526-8 978-3-319-74528-2},
  langid = {english},
}

@Article{OverviewEvaluationBluetooth,
  author = {Gomez, Carles and Oller, Joaquim and Paradells, Josep},
  title = {Overview and Evaluation of {Bluetooth Low Energy}: An Emerging Low-Power Wireless Technology},
  journal = {Sensors},
  volume = {12},
  year = {2012},
  number = {9},
  pages = {11734--11753},
  url = {https://www.mdpi.com/1424-8220/12/9/11734},
  pubmedid = {23112680},
  issn = {1424-8220},
  doi = {10.3390/s120911734}
}

@inproceedings{sahinSwarmRoboticsSources2005,
  title = {Swarm {{Robotics}}: {{From Sources}} of {{Inspiration}} to {{Domains}} of {{Application}}},
  booktitle = {Swarm {{Robotics}}},
  author = {{\c S}ahin, Erol},
  editor = {{\c S}ahin, Erol and Spears, William M.},
  year = 2005,
  pages = {10--20},
  publisher = {Springer},
  address = {Berlin, Heidelberg},
  doi = {10.1007/978-3-540-30552-1_2},
  isbn = {978-3-540-30552-1}
}

@article{zhangDynamicPrioritizedTask2025,
  title = {Dynamic and Prioritized Task Scheduling of Heterogeneous Multi-Robot Systems Using Deep Reinforcement Learning},
  author = {Zhang, Jiabing and Jia, Qingxuan and Zhang, Shiyu and Chen, Gang},
  year = 2025,
  month = jul,
  journal = {Neurocomputing},
  volume = {638},
  pages = {130184},
  issn = {0925-2312},
  doi = {10.1016/j.neucom.2025.130184}
}

@article{brambilla2013swarm,
  title = {Swarm robotics: a review from the swarm engineering perspective},
  author = {Brambilla, Manuele and Ferrante, Eliseo and Birattari, Mauro and Dorigo, Marco},
  journal = {Swarm Intelligence},
  volume = {7},
  pages = {1--41},
  year = {2013},
  doi = {10.1007/s11721-012-0075-2},
  publisher = {Springer}
}

@article{superiot,
  doi = {10.1088/2515-7647/ad1c6a},
  url = {https://doi.org/10.1088/2515-7647/ad1c6a},
  year = {2024},
  month = {jan},
  publisher = {IOP Publishing},
  volume = {6},
  number = {1},
  pages = {011001},
  author = {Katz, Marcos and Paso, Tuomas and Mikhaylov, Konstantin and Pessoa, Luis and Fontes, Helder and Hakola, Liisa and Leppäniemi, Jaakko and Carlos, Emanuel and Dolmans, Guido and Rufo, Julio and Drzewiecki, Marcin and Sallouha, Hazem and Napier, Bruce and Branquinho, André and Eder, Kerstin},
  title = {Towards truly sustainable {IoT} systems: {T}he {SUPERIOT} project},
  journal = {Journal of Physics: Photonics}
}

@inproceedings{aguzzi2024engineering,
  title = {Engineering Distributed Collective Intelligence in Cyber-Physical Swarms},
  author = {Aguzzi, Gianluca and Savaglio, Claudio},
  booktitle = {2024 20th International Conference on Distributed Computing in Smart Systems and the Internet of Things (DCOSS-IoT)},
  pages = {570--575},
  year = {2024},
  organization = {IEEE}
}

@article{khan2014design,
  title = {Design of a reconfigurable {RFID} sensing tag as a generic sensing platform toward the future Internet of Things},
  author = {Khan, Muhammad S and Islam, Mohammad S and Deng, Hai},
  journal = {IEEE Internet of Things Journal},
  volume = {1},
  number = {4},
  pages = {300--310},
  year = {2014},
  doi = {10.1109/JIOT.2014.2329189},
  publisher = {IEEE}
}

@article{jones2022dots,
  title = {{DOTS}: An open testbed for industrial swarm robotic solutions},
  author = {Jones, Simon and Milner, Emma and Sooriyabandara, Mahesh and Hauert, Sabine},
  journal = {arXiv preprint arXiv:2203.13809},
  year = {2022}
}

@article{anuraj2024dynamic,
  title = {Dynamic Swarm Orchestration and Semantics in {IoT} Edge Devices: A Systematic Literature Review},
  author = {Anuraj, Banani and Calvaresi, Davide and Aerts, Jean-Marie and Calbimonte, Jean-Paul},
  journal = {IEEE Access},
  volume = {12},
  pages = {116917--116938},
  year = {2024},
  doi = {10.1109/ACCESS.2024.3446876},
  publisher = {IEEE}
}

@article{kang2024optimization,
  title = {Optimization of Task Allocation for Resource-Constrained Swarm Robots},
  author = {Kang, Woosuk and Jeong, Eunjin and Shim, Sungjun and Ha, Soonhoi},
  journal = {IEEE Transactions on Automation Science and Engineering},
  volume = {22},
  pages = {3068--3085},
  year = {2025},
  doi = {10.1109/TASE.2024.3389013},
  publisher = {IEEE}
}

@article{kazim2025design,
  title = {Design and Deployment of Swarm Engineering Systems: Insights, Challenges, and Innovations},
  author = {Kazim, Raza Muhammad and Wang, Guoxin and Ming, Zhenjun and Allen, Janet K and Mistree, Farrokh},
  journal = {IEEE Access},
  volume = {13},
  pages = {139345--139376},
  year = {2025},
  doi = {10.1109/ACCESS.2025.3592950},
  publisher = {IEEE}
}

@incollection{dhaliwal2020rise,
  title = {The rise of automation and robotics in warehouse management},
  author = {Dhaliwal, Amandeep},
  booktitle = {Transforming management using artificial intelligence techniques},
  pages = {63--72},
  year = {2020},
  doi = {10.1201/9781003032410-5},
  publisher = {CRC Press}
}

@article{zhen2025optimizing,
  title = {Optimizing Warehouse Operations with Autonomous Mobile Robots},
  author = {Zhen, Lu and Tan, Zheyi and de Koster, Ren{\'e} and He, Xueting and Wang, Shuaian and Wang, Huiwen},
  journal = {Transportation Science},
  volume = {59},
  number = {5},
  pages = {1130--1152},
  year = {2025},
  doi = {10.1287/trsc.2024.0800},
  publisher = {INFORMS}
}

@inproceedings{ikumapayi2024swarm,
  title = {Swarm Robotics in a Sustainable Warehouse Automation: Opportunities, Challenges and Solutions},
  author = {Ikumapayi, Omolayo Michael and Laseinde, Opeyeolu Timothy and Elewa, Remilekun R and Ogedengbe, Temitayo Samson and Akinlabi, Esther Titilayo},
  booktitle = {E3S Web of Conferences},
  volume = {552},
  pages = {01080},
  year = {2024},
  doi = {10.1051/e3sconf/202455201080},
  organization = {EDP Sciences}
}

@inproceedings{notomista2019optimal,
  title = {An optimal task allocation strategy for heterogeneous multi-robot systems},
  author = {Notomista, Gennaro and Mayya, Siddharth and Hutchinson, Seth and Egerstedt, Magnus},
  booktitle = {2019 18th European control conference (ECC)},
  pages = {2071--2076},
  year = {2019},
  doi = {10.23919/ECC.2019.8795895},
  organization = {IEEE}
}

@article{szyc2023bluetooth,
  title = {Bluetooth low energy indoor localization for large industrial areas and limited infrastructure},
  author = {Szyc, Kamil and Nikodem, Maciej and Zdunek, Micha{\l}},
  journal = {Ad Hoc Networks},
  volume = {139},
  pages = {103024},
  year = {2023},
  doi = {10.1016/j.adhoc.2022.103024},
  publisher = {Elsevier}
}

@article{na2021bio,
  title = {Bio-inspired artificial pheromone system for swarm robotics applications},
  author = {Na, Seongin and Qiu, Yiping and Turgut, Ali E and Ulrich, Ji{\v{r}}{\'\i} and Krajn{\'\i}k, Tom{\'a}{\v{s}} and Yue, Shigang and Lennox, Barry and Arvin, Farshad},
  journal = {Adaptive Behavior},
  volume = {29},
  number = {4},
  pages = {395--415},
  year = {2021},
  publisher = {SAGE Publications Sage UK: London, England}
}

@article{jones2023frappe,
  title = {Frapp{\'e}: Fast Fiducial Detection on Low Cost Hardware},
  author = {Jones, Simon and Hauert, Sabine},
  journal = {Journal of Real-Time Image Processing},
  volume = {20},
  number = {6},
  pages = {119},
  year = {2023},
  doi = {10.1007/s11554-023-01373-w}
}

@article{jones2020distributed,
  title = {Distributed Situational Awareness in Robot Swarms},
  author = {Jones, Simon and Milner, Emma and Sooriyabandara, Mahesh and Hauert, Sabine},
  journal = {Advanced Intelligent Systems},
  volume = {2},
  number = {11},
  pages = {2000110},
  year = {2020},
  doi = {10.1002/aisy.202000110}
}

@inproceedings{hickson2025beesics,
  title = {Back to Bee-sics: Learning Information Sharing Strategies for Robot Swarms Through the Hive},
  author = {Hickson, Henry and Hauert, Sabine and Mavromatis, Alex},
  booktitle = {ALIFE 2025: Ciphers of Life: Proceedings of the Artificial Life Conference 2025},
  volume = {37},
  pages = {78},
  publisher = {MIT Press},
  year = {2025},
  doi = {10.1162/isal.a.906}
}

@article{azadeh2019robotized,
  title = {Robotized and Automated Warehouse Systems: Review and Recent Developments},
  author = {Azadeh, Kaveh and De Koster, Ren{\'e} and Roy, Debjit},
  journal = {Transportation Science},
  volume = {53},
  number = {4},
  pages = {917--945},
  year = {2019},
  doi = {10.1287/trsc.2018.0873}
}

\end{document}